\documentclass[letterpaper, 10 pt, conference]{ieeeconf} 
\IEEEoverridecommandlockouts                             
\usepackage{caption}
\usepackage[skins,most]{tcolorbox}  
\usepackage{subcaption}
\usepackage{multicol}
\usepackage{stfloats}
\usepackage{amsmath}
\usepackage{xcolor}
\let\labelindent\relax
\usepackage{enumitem}
\usepackage{siunitx}
\newcommand{\rn}[1]{\textit{#1)}}
 \usepackage{booktabs}
 \usepackage{float}
 \usepackage{algorithm}
\usepackage{algpseudocode}
\usepackage{amssymb}
\usepackage{mathtools}
\usepackage{graphics} 
\usepackage{epsfig} 
\usepackage{mathptmx}
\usepackage{times} 
\usepackage{amsmath} 
\usepackage{amssymb} 
\usepackage{graphicx}
\usepackage{epstopdf}
\usepackage{pdfpages}
\usepackage{titlesec}
\titleformat{\paragraph}[runin]{\normalfont\bfseries}{}{0pt}{}
\titlespacing*{\paragraph}{0pt}{1.5ex plus 1ex minus .2ex}{0.5em}
\usepackage{etoolbox}
\usepackage{amsbsy}
\usepackage{bm}
\usepackage{url}
\usepackage{accents}

\pdfoutput=1
\usepackage{hyperref}
\hypersetup{
    colorlinks=true,
    linkcolor=black,
    citecolor=black,
    filecolor=black,
    urlcolor=black,
}

\usepackage{tikz}
\usepackage[most]{tcolorbox}
\usepackage{cite}

\usepackage{stfloats}  
\def\horizontaldistance{\kern2pt}

\title{\LARGE \bf
Adaptive-MHE : A Sampling-Based Adaptive MPC for Legged Loco-Manipulation via Moving Horizon Estimation}
\author{Hossein Keshavarz$^{1}$, Alejandro Ramirez-Serrano$^{1}$, Majid Khadiv$^{2}$ 
\thanks{$^{1}$University of Calgary, Department of Mechanical Engineering, 2500 University Drive NW, Calgary, AB, Canada
Emails: {\tt\small {hossein.keshavarz},   {aramirez@ucalgary.ca}}}
\thanks{$^{2}$Munich Institute of Robotics and Machine Intelligence (MIRMI), Technical University of Munich (TUM), Germany
Email: {\tt\small {majid.khadiv@tum.de}}}
}
\makeatletter
\newcommand*{\rom}[1]{\expandafter\@slowromancap\romannumeral #1@}
\makeatother
\begin{document}

\maketitle
\thispagestyle{empty}
\pagestyle{empty}
%%%%%%%%%%%%%%%%%%%%%%%%%%%%%%%%%%%%%%%%%%%%%%%%%%%%%%%%%%%%%%%%%%%%%%%%%%%%%%%%
\begin{abstract}
Legged robots have demonstrated a remarkable ability to traverse various terrains, yet generating effective loco-manipulation behaviors remains challenging. A key difficulty is that object and terrain parameters are typically unknown to the robot, and mismatches between these parameters and their simulated counterparts introduce a sim-to-real gap that degrades control performance. Classical system identification (Sys-ID) methods often assume differentiable dynamics, an assumption that does not hold for contact-rich legged systems. Sampling-based Sys-ID avoids this restriction by directly matching simulated and recorded state trajectories through massively parallel rollouts, but existing approaches are typically applied offline and do not adapt as environmental conditions change. We present \emph{Adaptive-MHE} an online sampling-based Sys-ID framework, based on moving horizon estimation (MHE), that estimates the physical parameters of objects and terrain in the environment (e.g., mass, friction) and couples this estimate with a sampling-based model predictive controller, enabling adaptive loco-manipulation in changing and uncertain environments. In simulation and hardware experiments, our framework consistently outperforms baselines and matches the performance of a controller with access to ground-truth parameters.
\end{abstract}
%%%%%%%%%%%%%%%%%%%%%%%%%%%%%%%%%%%%%%%%%%%%%%%%%%
\section{Introduction}
\label{sec:introduction}
Humans and animals exploit a diverse set of body parts to interact with the environment in flexible ways, enabling them to traverse cluttered terrain and manipulate objects far larger than themselves. Replicating this level of contact-rich interactions in robots remains a major challenge. A key difficulty is that these interactions depend on uncertain physical parameters, including object mass, friction, and terrain properties, which are rarely known in advance and can vary substantially across tasks. As a result, fixed-parameter controllers often fail under uncertainty and environmental changes, leading to slipping, instability, or ineffective force application. Existing loco-manipulation methods often assume access to accurate prior knowledge or training-time dynamics. This limitation motivates the need for robots that can infer relevant physical properties directly from interaction data, enabling more adaptive and generalizable loco-manipulation in unseen environments.
\begin{figure}[htbp]
	\centering
	\includegraphics[width=\linewidth, trim={3.0cm 2.5cm 3.8cm 3.0cm}, clip]{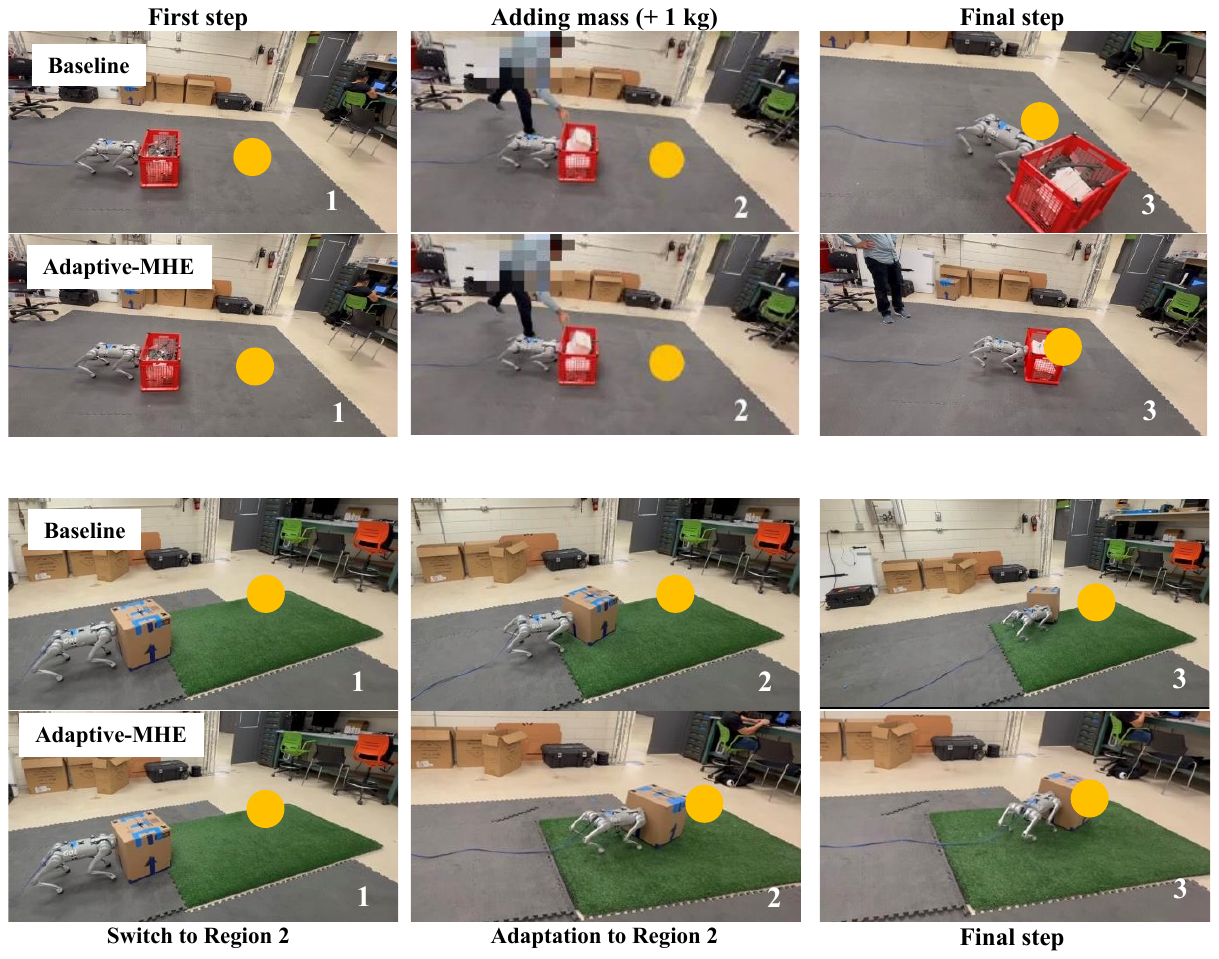}
	\caption{Comparison with the baseline: (top) \hyperref[task:T3]{T3}: Adaptation to mass variation, (bottom) \hyperref[task:T4]{T4}: Adaptation to terrain change.} 
	\label{fig:exp}
\end{figure}
In this work, we address the problem of loco-manipulation of unknown objects within a priori unknown environment parameters. To this end, we propose a sampling-based MHE framework that estimates unknown interaction parameters, including the object's mass and inertia as well as changes in the terrain friction coefficient, directly from the robot's most recent window of observations during interaction. Rather than solving the underlying nonlinear estimation problem via gradient-based optimization \cite{khorshidi2025physically,martinez2026system}, which requires derivatives of the dynamics under contact constraints, our formulation samples a large set of candidate parameters over the estimation horizon, evaluates each by comparing predicted and observed interaction trajectories, and selects the parameter set that best explains the robot's recent observed dynamics. Coupling this sampling-based MHE module with a sampling-based model predictive controller yields, to the best of our knowledge, the first sampling-based adaptive MPC formulation for legged loco-manipulation, in which parameter estimation and control are solved with a unified sampling-based framework. The resulting online estimates continually update the downstream controller, enabling robust object-manipulation and pushing behaviors across a wide range of object properties and terrain conditions.

The main contributions of this paper are as follows:
\begin{itemize}
    \item We formulate a novel sampling-based moving horizon estimation (MHE) mechanism for online system identification in legged loco-manipulation, efficiently identifying the key physical parameters that most affect controller performance from a sliding window of real-world interaction data, without requiring a differentiable simulator or specialized sensing (e.g., touch or force/torque sensors).
    \item We couple this sampling-based MHE module with a sampling-based model predictive controller, yielding, to the best of our knowledge, the first sampling-based adaptive MPC formulation for legged loco-manipulation, and demonstrate improved performance under object and terrain uncertainty across simulation and hardware experiments.
\end{itemize}

\section{Related Work}
\label{sec:background}
\subsection{Loco-Manipulation for quadruped Robots}
Learning-based approaches to loco-manipulation control typically rely on reinforcement learning (RL) to train whole-body policies and have demonstrated impressive generalization across a variety of tasks~\cite{jeon2023learning,pan2025roboduet,omar2025learning,dadiotis2025dynamic} (see \cite{ha2025learning} for a complete review of different methods). These policies are typically trained over a fixed distribution of object and terrain properties through domain randomization, and adapting to parameters outside this distribution generally requires retraining or fine-tuning. To address this limitation, object and terrain parameters can instead be estimated online and incorporated into the control loop, enabling adaptation to new conditions without retraining the underlying policy.

In the model-based control community, loco-manipulation has traditionally been formulated as an optimization problem and solved using gradient-based methods~\cite{wensing2023optimization}. These approaches generally assume that the relevant robot and environment parameters are known in advance. While some works have attempted to estimate unknown parameters through system identification for contact-rich locomotion~\cite{khorshidi2025physically,martinez2026system}, such methods are often computationally expensive, limiting their applicability to online estimation. Consequently, tightly coupling these estimation methods with MPC to achieve real-time adaptive control remains challenging.

With recent advances in massively parallel simulation, sampling-based MPC has emerged as an attractive alternative for controlling robots in contact-rich settings. Unlike RL, it requires no offline policy training and can leverage parallel sampling to handle discontinuous and non-convex dynamics. Recent works have successfully applied sampling-based MPC to legged loco-manipulation~\cite{xue2025full,alvarez2025real,keshavarz2025control,pacelli2026sampling}. More recent approaches have also explored combining the complementary advantages of sampling-based MPC and RL~\cite{crestaz2025td,brudermuller2026generative,zhang2026sumo,shirwatkar2026shield}. However, these approaches generally assume that the relevant properties of the environment are known, limiting their ability to adapt online to changing or previously unknown environmental conditions.

\subsection{System Identification and Parameter Adaptation}
System identification for robotic systems dates back to the late 1980s~\cite{atkeson1986estimation}. With advances in constrained optimization, recent approaches have established the interplay between geometric and physical consistency in system identification by explicitly enforcing constraints~\cite{wensing2017linear,lee2024robot}. These optimization-based approaches typically require derivatives of the dynamics, limiting their applicability to contact-rich settings. Several recent works have sought to extend system identification to systems with contact switches~\cite{khorshidi2025physically,martinez2026system,kang2026prime}. However, these methods rely on precise contact-event detection, which limits their applicability to highly contact-rich scenarios such as whole-body loco-manipulation, where frequent and intermittent contacts occur between the robot and its environment.

More recently,~\cite{sobanbabu2025sampling} introduced Sampling-based Parameter Identification (SPI) with active exploration, a two-stage sampling-based framework that identifies a legged robot's physical parameters from real-world trajectories collected using motion priors from pre-trained RL policies. The framework subsequently refines the parameter estimates by actively selecting exploration commands that maximize the Fisher information~\cite{ly2017tutorial} of the collected data. However, SPI is applied offline, limiting its ability to adapt to changing conditions during operation. Also, it was only demonstrated for locomotion.
In contrast, we propose Adaptive-MHE, an online sampling-based parameter estimation framework that estimates the physical parameters of objects and the environment during loco-manipulation. By coupling Adaptive-MHE with sampling-based MPC, our approach enables adaptive control in the presence of unknown and changing object properties.

\section{Preliminaries}
\label{sec:Preliminaries}
This section briefly explains two building blocks underlying the proposed framework: MPOPI \cite{keshavarz2025control}, the sampling-based whole-body MPC controller, and SPI \cite{sobanbabu2025sampling}, the sampling-based parameter identification method that our online estimator builds upon.

\subsection{MPOPI}
MPOPI~\cite{keshavarz2025control} combines MPPI~\cite{alvarez2025real} with the Covariance Matrix Adaptation Evolution Strategy (CMA-ES) method~\cite{hansen2016cma} to leverage the advantages of each, with the goal of creating a unified approach that reduces the time needed to find a control action~\cite{asmar2023model}. In contrast to MPPI and CMA-ES individually, MPOPI generates an initial set of randomized samples using a random or predefined mean and covariance. These two parameters are then updated via CMA-ES, from which a new, improved set of randomized samples is generated using the updated mean and covariance. This updating process is repeated for $L$ cycles, yielding a refined subset of robot motion trajectories (joint positions). This strategy enables an efficient search over actions to quickly find the optimal sequence with reduced computation.

\subsection{SPI}
\label{sec:spi_background}
SPI~\cite{sobanbabu2025sampling} formulates system identification as a zero-order optimization problem solved via GPU-based parallel sampling, avoiding the need for differentiable dynamics. The system is modeled as $x_{t+1} = f(x_t, u_t; \theta)$, where $x_t$ is the robot state, $u_t$ the control input, and $\theta$ the unknown physical parameters to be identified from a dataset of real-world state-action trajectories $\mathcal{D} = \{(x_t, u_t)\}_{t=1}^{N}$. The identified parameters $\theta = [\theta_{in}, \theta_{mo}]^T$ consist of mass-inertia properties $\theta_{in}$ (mass, center of mass, and inertia) and actuator parameters $\theta_{mo}$ characterizing motor dynamics, whose parameterization ensures physical correctness of the outputs \cite{lee2019geometric}. Real-world trajectories are segmented into $N_c$ short clips of horizon length $H$, sampled from a uniform distribution to avoid bias from any single fixed horizon length. The identification problem is then posed as a multi-step, nonlinear least-squares rollout-matching objective
\begin{equation}
J(\theta, \{c_k\}) = \sum_{k=1}^{N_c} \sum_{t=0}^{H-1} \left\| x^r_{t+1,k} - x_{t+1,k} \right\|^2_{W_x} + \left\| \theta - \theta_0 \right\|^2_{W_\theta},
\label{eq:spi_cost}
\end{equation}
where $x_{t+1,k} = f(x_{t,k}, u^r_{t,k}; \theta)$ is the simulated state rolled out under candidate parameter $\theta$ from clip $c_k$'s recorded initial condition, while the second cost term regularizes $\theta$ toward a nominal prior $\theta_0$. This cost is minimized using CMA-ES: at each iteration, a batch of candidate parameter vectors is sampled, evaluated in parallel via~\eqref{eq:spi_cost}, and used to update the sampling distribution until convergence, yielding the parameter estimate $\hat\theta$.

\section{Proposed locomotion framework}
\label{sec:framework}
\begin{figure}[htbp]
	\centering
	\includegraphics[width=\linewidth, trim={5cm 0.5cm 2cm 0.5cm}, clip]{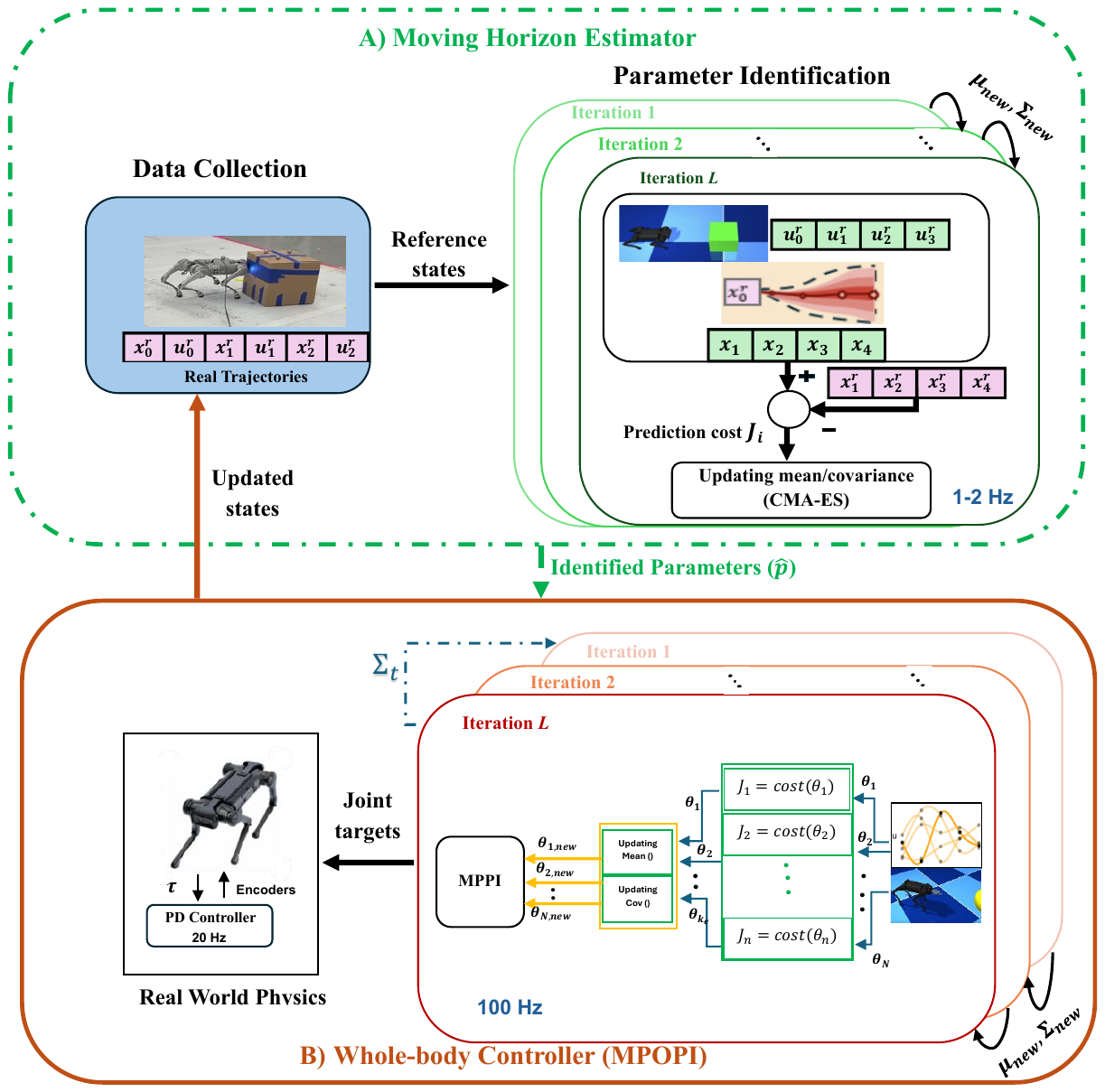}
	\caption{Overview of the Adaptive-MHE framework.} 
	\label{fig:Flowchart}
\end{figure}

The proposed adaptive control framework builds upon SPI~\cite{sobanbabu2025sampling} and MPOPI~\cite{keshavarz2025control}, extending SPI's sampling-based estimation strategy to an online moving-horizon estimation (MHE) framework that recursively re-estimates object and terrain parameters from interaction data collected during task execution. The framework is illustrated in Fig.~\ref{fig:Flowchart}, consists of two main components: \rn{A} the moving-horizon estimator and \rn{B} the whole-body controller.
The two components operate asynchronously on separate CPU threads, with the estimator running in the background at a lower frequency while the whole-body controller continues to generate control actions in real time.

\subsection{Moving Horizon Estimator}
\label{sec:mhe}

SPI~\cite{sobanbabu2025sampling} identifies system parameters offline from a fixed, pre-collected dataset using massively parallel GPU sampling. In contrast, the proposed estimator operates online and continuously updates its parameter estimate as new interaction data become available. During task execution, the states of both the robot and the box, together with the corresponding control inputs applied to the robot, are recorded at each time step to form the trajectory
\begin{equation}
    \mathcal{D}
    =
    \left\{(x_t^r,u_t)\right\}_{t=1}^{N},
    \label{eq:ref_state}
\end{equation}
where $x_t^r$ contains the pose (position and orientation), linear and angular velocities of the robot base and box, as well as the robot joint positions $q$ and velocities $\dot{q}$.

Contrary to SPI that estimates the parameters once from the complete recorded history, our proposed estimator uses a short sliding window containing only the most recent history of interaction data. At each update, the window
\begin{equation}
    \mathcal{D}_H(t)
    =
    \left\{(x_\tau^r,u_\tau)\right\}_{\tau=t-H+1}^{t}
\end{equation}
is extracted from $\mathcal{D}$ and used for parameter estimation. The window is refreshed every $\Delta t_{\mathrm{update}}$ seconds, allowing the estimate to be refined as additional information about the system dynamics becomes available.

Unlike SPI, which parallelizes candidate rollouts on a GPU, the proposed estimator runs on the CPU in a separate thread from the whole-body controller. This asynchronous implementation allows parameter estimation to proceed in the background without interrupting the real-time control loop.

\subsection{Whole-body Controller (MPOPI)}

The estimated parameters are incorporated into the dynamics model used by the MPOPI controller introduced in Section~\ref{sec:Preliminaries}. In~\cite{keshavarz2025control}, the sampling distribution is refined over $L_{MPOPI}$ loops at every control cycle, which increases the computational cost of the controller. To reduce this cost, we exploit the periodic structure of the reference gait and perform the multi-loop refinement only when the contact mode changes. Between mode switches, the sampling distribution obtained from the most recent refinement is retained and used to warm-start the subsequent control cycles.
This modification reduces the computational burden of MPOPI while preserving its sampling-based exploration and exploitation during complex whole-body locomotion tasks. The resulting controller operates continuously while the MHE thread asynchronously updates the dynamics parameters.

\subsection{Adaptive-MHE Formulation}
\label{sec:final_formulation}

Algorithm~\ref{alg:adaptive_mhe} summarizes the complete control loop. Both the estimator and the controller use the same parameterized dynamics model $f$. Its state transition is given by
\begin{equation}
    \mathbf{x}_{t+1}
    =
    f(\mathbf{x}_t,\mathbf{u}_t;p)
    +
    \mathbf{w}_t,
    \label{eq:final_dynamics}
\end{equation}
where $\mathbf{x}_t$ and $\mathbf{u}_t$ denote the system state and control input, respectively, $\mathbf{w}_t$ represents process noise, and $p$ is the vector of unknown system parameters. For brevity, $f(p)$ denotes the dynamics model instantiated with parameter vector $p$.

At each control cycle, MPOPI (line 13-22 in  Algorithm~\ref{alg:adaptive_mhe}) uses the current estimate $\hat{p}$ to generate and evaluate candidate control sequences using $f(\hat{p})$ (line 4 in  Algorithm~\ref{alg:adaptive_mhe}). The first control action of the optimized sequence is applied to the system, and the resulting state--input pair is appended to the data buffer. When the parameter-update interval $\Delta t_{\mathrm{update}}$ is reached, the most recent $H$ measurements are used to update the parameter estimate.

Given the current identification window and the previous parameter estimate $\hat{p}_t^{-}$, the MHE problem is formulated as
\begin{align}
    \hat{p}_t =
    \arg\min_{p} \sum_{k=0}^{H-1}
    \left\|
        x^r_{k+1}-x_{k+1}
    \right\|_{W_x}^{2}
    +
    \left\|
        p-\hat{p}_t^{-}
    \right\|_{W_p}^{2}, 
    \label{eq:final_mhe}
\end{align}

Here, $x^r_{k+1}$ denotes the recorded state in the identification window, while $x_{k+1}$ is the corresponding state predicted by rolling out $f(p)$ using the recorded control inputs. The matrix $W_x$ weights the state prediction error, while $W_p$ regularizes the updated estimate toward the previous estimate. In this work, $W_p =\mathrm{diag}    \left(w_{\mathrm{mass}}, w_{\mathrm{inertia}}, w_{\mathrm{friction}}\right).$

To solve the optimization in~\eqref{eq:final_mhe}, a diagonal Gaussian distribution is initialized around the previous estimate $\hat{p}_t^{-}$ and iteratively refined using candidate parameter samples (line 23-24 in  Algorithm~\ref{alg:adaptive_mhe}). At each iteration $\ell$, $B$ candidates are drawn according to
\begin{equation}
    p_j^{(\ell)}
    \sim
    \mathcal{N}\!\left(\mu_p^{(\ell)},\Sigma_p^{(\ell)}\right),
    \qquad
    \Sigma_p^{(\ell)}
    =
    \operatorname{diag}\!\left((\sigma_p^{(\ell)})^2\right).
\end{equation}
Each candidate is evaluated by rolling out $f(p_j^{(\ell)})$ over the identification window and the candidate distribution is then updated from the lowest-cost samples using log-rank-weighted averaging with learning rate $\alpha$. The resulting distribution mean provides the updated estimate $\hat{p}_t$.

The updated estimate is subsequently used to instantiate $f(\hat{p}_t)$ for the controller (line 9 in  Algorithm~\ref{alg:adaptive_mhe}). As new measurements are collected, the identification window advances and the estimation procedure is repeated every $\Delta t_{\mathrm{update}}$ seconds. Thus, parameter estimation and whole-body control proceed concurrently, allowing the controller to adapt its dynamics model online while maintaining the real-time control loop. The complete procedure is summarized in Algorithm~\ref{alg:adaptive_mhe}.

\begin{algorithm}[htbp]
\caption{\textsc{\textcolor{green!50!black}{Adaptive-MHE}} framework}
\label{alg:adaptive_mhe}
\begin{algorithmic}[1]
\Require Initial parameter prior $p_0$, dynamics model $f(p)$, estimator hyperparameters
$(L_{\mathrm{MHE}}, B, \Delta t_{\mathrm{update}}, H, \alpha, w_{\mathrm{mass}}, w_{\mathrm{inertia}}, w_{\mathrm{friction}})$
\Ensure Control actions $\{u_t\}$; updated parameter estimate $\hat{p}$
\State Initialize $\hat{p} \leftarrow p_0$, data buffer $\mathcal{D} \leftarrow \emptyset$
\State Initialize trajectory mean $\mu_t$ and covariance $\Sigma_t$
\For{$t \leftarrow 0$ to $T-1$}
    \State $(u_t,\mu_t,\Sigma_t)
    \leftarrow
    \textsc{\textcolor{blue!60!black}{MPOPI}}(\mu_t,\Sigma_t,\hat{p})$
    \State Append $(x_t,u_t)$ to $\mathcal{D}$
    \If{parameter update is triggered}
        \State $\mathcal{D}_H
        \leftarrow
        \left\{(x_\tau,u_\tau)\right\}_{\tau=t-H+1}^{t}$
        \State $\hat{p}
        \leftarrow
        \textsc{\textcolor{red!60!black}{MHE-Estimate}}(\mathcal{D}_H,\hat{p}^-)$
        \State Update dynamics model $f(\hat{p})$
    \EndIf
\EndFor
\State \Return $\{u_t\}$, $\hat{p}$

\Function{\textcolor{blue!60!black}{MPOPI}}{$\mu_t,\Sigma_t,\hat{p}$}
    \State Sample and parallel-roll out $N$ action sequences under $f(\hat{p})$; evaluate their costs
    \If{gait mode switch}
        \State Refine $\Sigma_t$ over $L_{\mathrm{MPOPI}}$ loops
    \Else
        \State Reuse $\Sigma_t$ as a warm start
    \EndIf
    \State Update $\mu_t$ using weighted costs; $u_t \leftarrow \mu_t[0]$
    \State \Return $u_t,\mu_t,\Sigma_t$
\EndFunction

\Function{\textcolor{red!60!black}{MHE-Estimate}}{$\mathcal{D}_H,\hat{p}$}
    \State Initialize the parameter distribution with mean $\hat{p}$
    \For{$\ell \leftarrow 1$ to $L_{\mathrm{MHE}}$}
        \State Sample $\{p_j\}_{j=1}^{B}
        \sim
        \mathcal{N}(\mu_p,\mathrm{diag}(\sigma_p^2))$
        \State Evaluate trajectory cost $J(p_j)$ using~\eqref{eq:final_mhe}
        \State Update $\mu_p,\sigma_p$ using $\{J(p_j)\}_{j=1}^{B}$
        \If{$\sigma_p \leq \sigma_{\mathrm{floor}}$}
            \State \textbf{break}
        \EndIf
    \EndFor
    \State \Return $\hat{p} \leftarrow \mu_p$
\EndFunction
\end{algorithmic}
\end{algorithm}

\section{Evaluation}  
\label{sec:evaluation}
In this section, we evaluate the proposed framework on several loco-manipulation tasks using the Unitree Go1. The controller is implemented in Python within the Robot Operating System (ROS) framework and runs in real time on an AMD Ryzen 9 7950X3D CPU with 64~GB of memory. Our objective is to demonstrate that the robot can successfully manipulate objects while adapting to changes in the terrain and the physical properties of the object. Throughout all tasks, we compare our method \textit{Adaptive-MHE} to two alternatives, \emph{True}, and \emph{Baseline}. The \emph{True} control mode uses a fixed belief based on the ground-truth parameters, representing the best achievable performance with a non-adaptive belief. In contrast, \textit{Adaptive-MHE} first initializes its belief via an offline probing estimation step and subsequently performs online closed-loop re-identification during task execution using a moving horizon estimator. Finally, \emph{Baseline} represents the SPI \cite{sobanbabu2025sampling} that relies on a fixed, offline-identified prior belief that remains unchanged throughout the run, reflecting the standard practice of offline parameter identification without further adaptation.

\subsection{Task Definitions and setups}
\label{sec:task_defs}
\begin{figure}[htbp]
\centering
\includegraphics[width=.99\linewidth, trim={3cm 8.5cm 3cm 7.0cm}, clip]{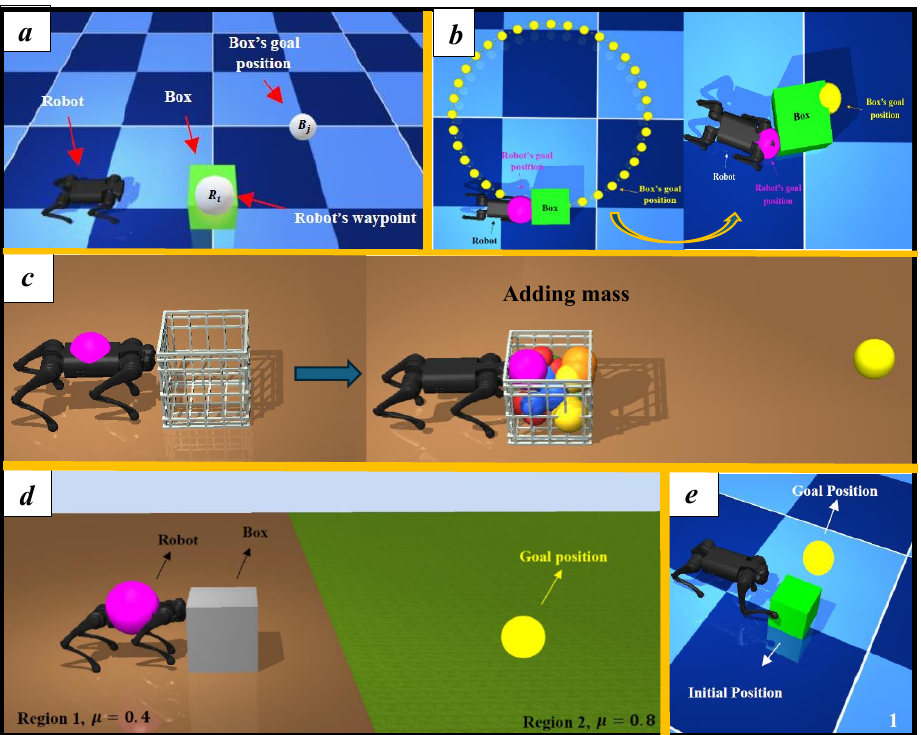}
\caption{Overview of the loco-manipulation scenarios evaluated in this work: (a) T1. Single-goal pushing, (b) T2. Circular-path pushing, (c) T3. Adaptation to mass variation, (d) T4. Adaptation to terrain uncertainty, and (e) T5. Pushing a box with one leg.}
\label{fig:tasks}  
\end{figure}

We design a series of simulation scenarios covering diverse loco-manipulation tasks. For comparison across control modes, the robot is commanded to move at $0.2$~m/s in the following tasks:

\paragraph{T1. Single-goal pushing:} 
\label{task:T1}
The robot pushes an object with unknown physical properties to a predefined goal. As shown in Fig.~\ref{fig:tasks}a, the box waypoint ($B_j$) remains fixed at the goal, while the robot waypoint ($R_i$) tracks the box center to maintain contact during motion.

\paragraph{T2. Circular-path pushing:}
\label{task:T2}
To evaluate continuous adaptation, the robot pushes an unknown box along a circular trajectory (Fig.~\ref{fig:tasks}b). The robot waypoint is placed on the pushing side of the box and continuously updated toward the next target position, ensuring the correct pushing direction along the curved path.

\paragraph{T3. Adaptation to mass variation:}
\label{task:T3}
The box mass is changed abruptly during the task (Fig.~\ref{fig:tasks}c), requiring \textit{Adaptive-MHE} to re-estimate the mass and associated friction online and adapt the control action accordingly.

\paragraph{T4. Adaptation to terrain change:}
\label{task:T4}
The robot pushes the box across two floor regions with friction coefficients $\mu_1=0.4$ and $\mu_2=0.8$, separated at $x_b=1.5$~m (Fig.~\ref{fig:tasks}d). This task evaluates the ability to detect changes in terrain friction online and adapt the control action accordingly.

\paragraph{T5. Pushing a box with one leg:} 
\label{task:T5}
The robot pushes a box of unknown physical properties $0.5$~m to the side using a front leg while maintaining a stationary three-legged stance (Fig.~\ref{fig:tasks}e). Unlike the previous tasks, the desired robot speed is zero.

\subsection{Ablation and Simulation Study}
\label{sec:ablation}
We evaluate two aspects of the proposed online \textit{Adaptive-MHE} pipeline: (i) how sensitive closed-loop task performance is to the online identification hyperparameters, and (ii) whether \textit{Adaptive-MHE} provides a measurable benefit over a fixed belief under a controlled, non-stationary physical parameter change. All experiments use the box-pushing loco-manipulation task, in which a quadruped pushes a box toward a fixed goal while the floor friction changes (Fig.~\ref{fig:tasks}d).

\begin{figure*}[t]
\centering
\includegraphics[width=\linewidth, trim={0.0cm 0.0cm 0.0cm 0.0cm}, clip]{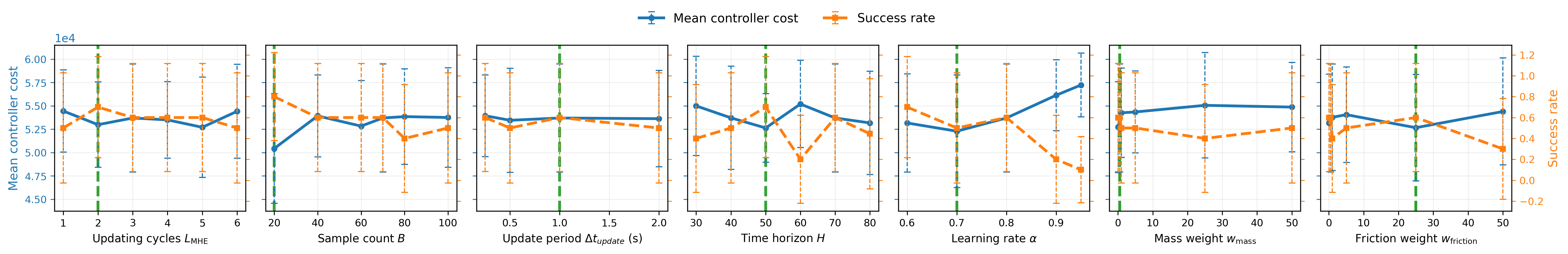}\vspace{-3mm}
\caption{Ablation over \textit{Adaptive-MHE} hyperparameters. The means and standard deviations are computed from 15 different random seeds for each setting. Green dashed lines denote the default parameters we deploy on hardware.}
\label{fig:ablation}
\end{figure*}

The proposed framework exposes seven hyperparameters in~\eqref{eq:final_mhe} that govern the accuracy and responsiveness of the estimator: the number of updating cycles $L_{MHE}$, sample count $B$, update period $\Delta t_\text{update}$ (seconds between identification calls), time horizon $H$, learning rate $\alpha$, and the regularization weights $w_\text{mass}$ and $w_\text{friction}$ that pull each online estimate back toward its warm-started prior. For each run, final goal error ($e_g$), controller cost ($\mathbf{J}$), success rate ($\mathcal{S}_R$), and execution time ($t_{\text{exec}}$) are reported. We vary each hyperparameter individually across a small grid of values while holding the remaining six fixed, repeating every configuration across $N_\text{seed} = 15$ random seeds. Figure~\ref{fig:ablation} shows the resulting mean $\pm$ standard deviation (across seeds) for mean controller cost and success rate. For each hyperparameter, the default value deployed on hardware (green dashed line) was selected as the setting that achieved the lowest mean controller cost while maintaining a high success rate, favoring the most conservative (i.e., most stable) choice among comparable configurations.

Using the default operating point identified in Fig.~\ref{fig:ablation} (green dashed lines), we compare the three belief conditions (\emph{True}, \emph{Baseline}, and \emph{Adaptive-MHE}) across the same $N_\text{seed} = 15$ seeds per condition, enabling paired comparisons. Table~\ref{tab:ablation-main} summarizes the results. The Baseline belief succeeds on only a small fraction of seeds. Injecting the ground-truth parameter values without further adaptation in \emph{True} mode achieves complete success and the lowest final goal error. \textit{Adaptive-MHE} substantially improves all metrics over the \emph{Baseline} belief despite never having access to ground-truth parameters, though it does not fully close the gap to the \emph{True} condition.

\begin{table}[htbp]
    \centering
    \caption{Comparison across control modes, mean $\pm$ std
    over 15 random seeds.}
    \label{tab:ablation-main}
    \footnotesize
    \setlength{\tabcolsep}{3pt}
    \begin{tabular}{@{}lccc@{}}
        \toprule
         & True &  Baseline& Adaptive-MHE (ours) \\
        \midrule
        $\mathcal{S}_R\ \uparrow$         &  $100.0\pm 0.0$ &  $13.33\pm 35.2$  &  $86.7\pm 35.2$ \\
        $e_g\ [\text{m}]\downarrow$             &  $0.045 \pm 0.02$ & $0.58\pm 0.23 $  &  $0.16\pm0.24$   \\
        $\mathbf{J}\ (\times 10^5)\downarrow$   &  $0.43\pm0.03$   &  $0.62\pm0.04$    &  $0.51\pm0.05$    \\
        $t_{\text{exec}}\ [\text{s}]\downarrow$  &  $48.1\pm0.2$ &  $77.4\pm0.13$       &  $56.3\pm0.04$   \\
        \bottomrule
    \end{tabular}
\end{table}

% \subsection{Simulation Results}
% \label{sec:sim_validation}
We also evaluate the performance of the three control modes across all five tasks introduced in Section~\ref{sec:task_defs}. Table~\ref{tab:sim_summary} summarizes the final goal/tracking error and RMSE for all tasks. As shown, our \textit{Adaptive-MHE} approach nearly matches the tracking performance of the controller with true parameters by continuously estimating the unknown parameters and incorporating them into the control loop. A detailed qualitative analysis of the results is provided in the Appendix.

% we discuss the two tasks that most directly demonstrate online adaptation to non-stationary parameters (\hyperref[task:T3]{T3}, \hyperref[task:T4]{T4}) in detail below, along with the simplest case (\hyperref[task:T1]{T1}) and the single-leg embodiment (\hyperref[task:T5]{T5}). Full results for \hyperref[task:T2]{T2} (circular-path pushing) are provided in Appendix~\ref{app:additional_results}, as its conclusions parallel \hyperref[task:T1]{T1}'s.

\begin{table}[htbp]
    \centering
    \caption{Summary of final error / RMSE across all tasks, True / Baseline / Adaptive-MHE (ours).}
    \label{tab:sim_summary}
    \footnotesize
    \setlength{\tabcolsep}{3pt}
    \begin{tabular}{@{}lccc@{}}
        \toprule
        Task & True & Baseline & Adaptive-MHE \\
        \midrule
        T1 -- RMSE (m/s) & $0.134$ & $0.160$ & $0.139$ \\
        T2 -- RMSE (m/s) & $0.108$ & $0.119$ & $0.104$ \\
        T3 -- RMSE (m/s) & $0.093$ & $0.117$ & $0.092$\\
        T4 -- RMSE (m/s) & $0.095$ & $0.099$ & $0.097$ \\
        T5 -- final error (m) & $0.12$ & $0.42$ & $0.28$ \\
        T5 -- RMSE (m/s) & $0.051$ & $0.149$ & $0.052$ \\
        \bottomrule
    \end{tabular}
\end{table}

\subsection{Real-world demonstration}
\label{sec:experiment}
In addition to the simulation studies, real-world locomanipulation experiments were conducted on a Unitree Go1 quadruped robot running either \textit{Adaptive-MHE} or \emph{Baseline} policies performing a box-pushing task. The robot's objective was to push an object (a box or a basket) toward a predefined goal position, while the object's physical properties, surface friction or mass, were varied in ways unknown to the policy a priori. 

In the first experiment (Fig.~\ref{fig:exp} top), we tested adaptation to an abrupt, unmodeled change in object mass during task execution. The robot pushed a basket with an initial mass of $3$~kg toward the target ("First step"). Partway through the push, an additional $1.0$~kg was manually added to the basket ("Adding mass"), increasing its total mass to $4.0$~kg with no explicit signal to the policy. The \emph{Baseline} policy ("Final step" top) continues applying force calibrated to the original $3$~kg estimate; since this force is insufficient for the heavier basket, the robot fails to close the remaining distance, leaving a noticeable gap to the target marker. The adaptive policy ("Final step" bottom) instead detects the change in the basket's response, re-estimates the mass online, and increases its pushing force accordingly, successfully driving the basket to the target despite the mid-episode disturbance.

In the second experiment, we evaluated robustness to spatially varying surface friction using an environment with two regions: a low-friction concrete floor (Region~1) and a high-friction turf mat (Region~2), as shown at the bottom of Fig.~\ref{fig:exp}. The robot pushed a box starting on the low-friction surface across onto the high-friction region toward the target. The \emph{Baseline} policy applies force calibrated to an average friction coefficient learned during offline estimation; snapshots 2--3 show that once the box crosses onto the high-friction mat, this mismatch causes the robot to under-drive the push, stalling the box short of the target. In contrast, the \textit{Adaptive-MHE} mode infers the increased resistance from the contact dynamics and compensates by increasing the applied force, successfully driving the box to the target (snapshots 2--3). This demonstrates that the online-estimated friction parameter enables the policy to adapt its control effort in real time.

\section{Conclusions}
\label{section:6}
In this paper, we presented \textit{Adaptive-MHE}, a sampling-based moving horizon estimation framework for online system identification coupled with sampling-based MPC for legged loco-manipulation. To the best of our knowledge, this is the first sampling-based adaptive MPC framework for legged loco-manipulation that does not require differentiable dynamics or hand-crafted adaptation laws. Simulation and hardware experiments demonstrated accurate trajectory and speed tracking under uncertain object and terrain parameters, with performance comparable to the True mode and consistently better than the \emph{Baseline} control mode, achieving substantially lower final goal error. These results demonstrate the potential of \textit{Adaptive-MHE} for robust legged loco-manipulation in unstructured environments.

%%%%%%%%%%%%%%%%%%%%%%%%%%%%%%%%%%%%%%%%%%%%%%%%%%%%%%%%%%%%%%%%%%%%%%%%%%%%%%%%
\bibliography{bibliography}
\bibliographystyle{IEEEtran}

\appendix
\section{Additional Simulation Results}
\label{app:additional_results}

Here we present qualitative results to better understand where the performance gain comes from. 

\begin{figure}[htbp]
	\centering
    \begin{subfigure}[]{0.45\linewidth}
	\includegraphics[width=\linewidth, trim={3.2cm 8.0cm 3.5cm 9.4cm}, clip]{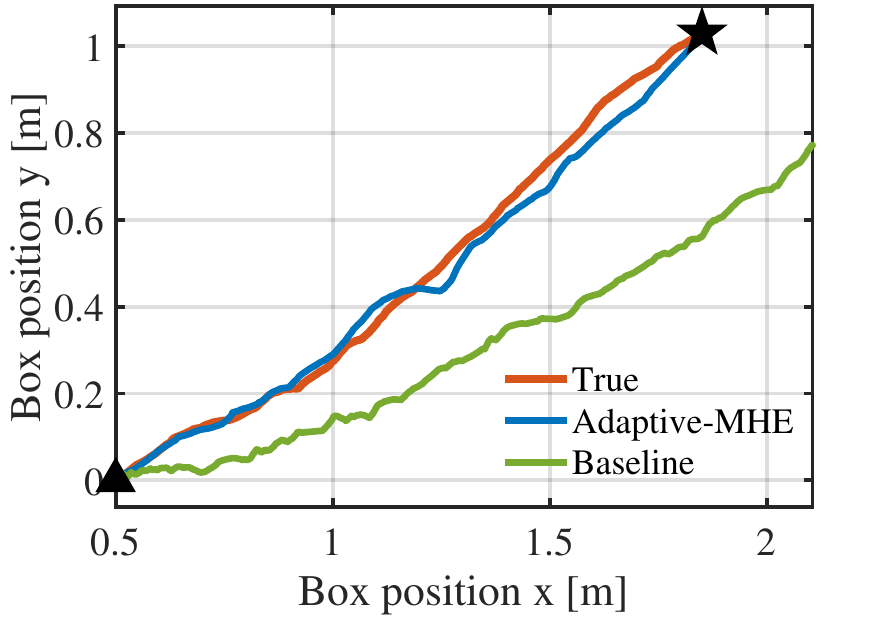}
    \label{fig:traj_point}
    \end{subfigure}%
    \hfill
    \begin{subfigure}[]{0.55\linewidth}
    \includegraphics[width=\linewidth, trim={0.0cm 7.0cm 0.0cm 7.0cm}, clip]{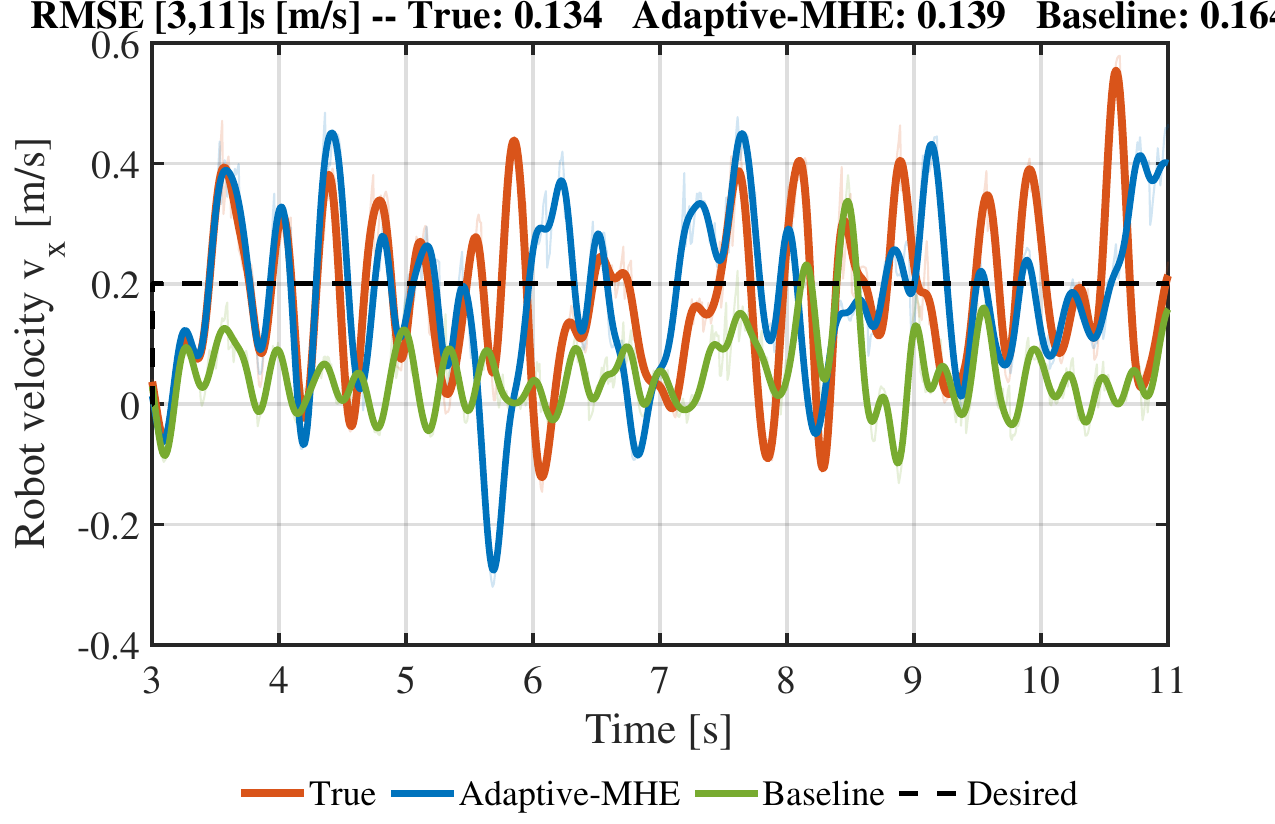}
    \label{fig:velocity_point}
    \end{subfigure}%
    \vspace{-6mm}
    \caption{Comparison of control modes for T1. (left) box trajectory; (right) robot forward velocity tracking.}
    \label{fig:T1}
\end{figure}

\paragraph{T1. Single-goal pushing:} We begin with the simplest case, i.e., pushing an object to the goal. Figure~\ref{fig:T1} compares the three control modes for this task. On the left, both \emph{True} and \emph{Adaptive-MHE} push the box to the goal successfully, while \emph{Baseline} deviates substantially. On the right, \emph{True} and \emph{Adaptive-MHE} track the commanded reference $v_x = 0.2$~m/s more closely than \emph{Baseline}, yielding lower RMSE (Table~\ref{tab:sim_summary}). These results confirm that inaccurate dynamics parameters degrade velocity tracking, and \emph{Adaptive-MHE} recovers performance close to \emph{True}.

\begin{figure}[htbp]
	\centering
    \begin{subfigure}[]{0.42\linewidth}
	\includegraphics[width=\linewidth, trim={2.0cm 7.0cm 3.5cm 7.5cm}, clip]{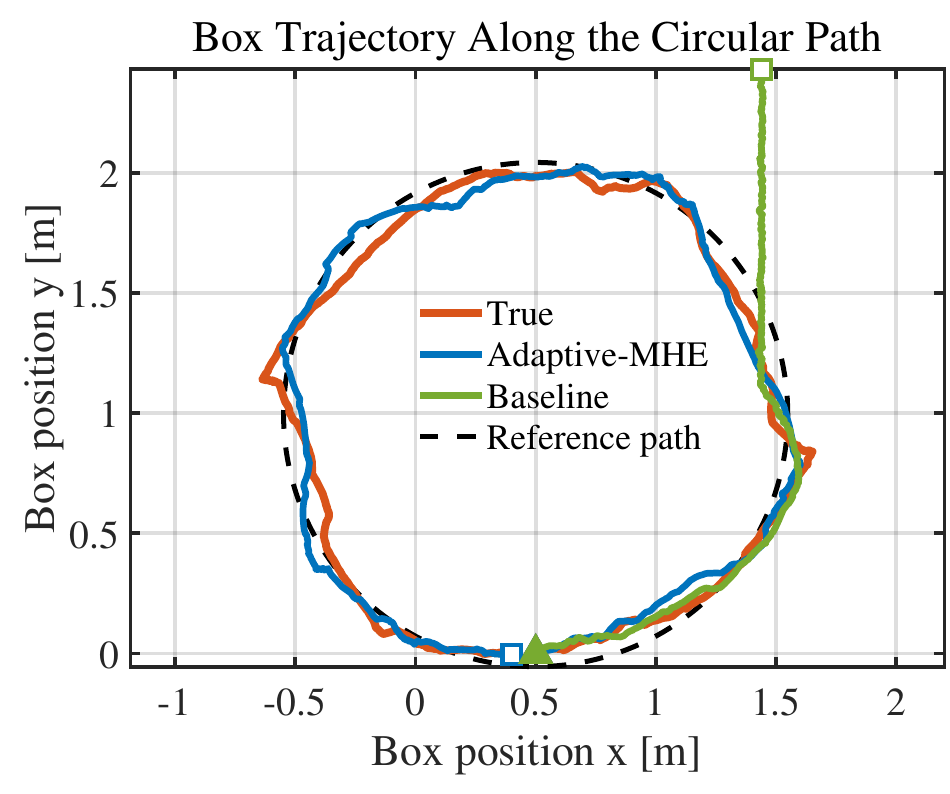}
	\label{fig:circular_traj}
    \end{subfigure}%
    \hfill
    \begin{subfigure}[]{0.58\linewidth}
    \includegraphics[width=\linewidth, trim={0.0cm 7.0cm 0.0cm 7.0cm}, clip]{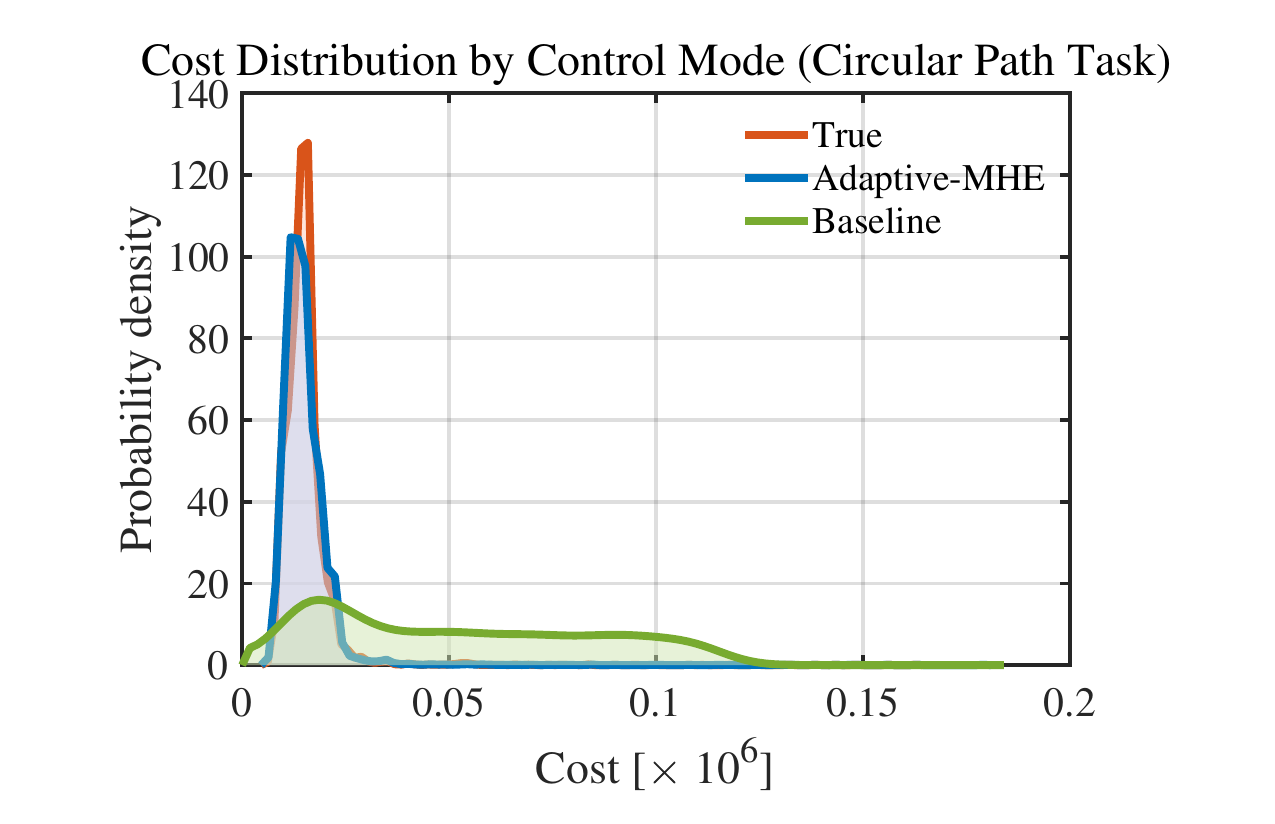}
	\label{fig:circular_cost}
    \end{subfigure}%
    \vspace{-6mm}
    \caption{Comparison of control modes for \hyperref[task:T2]{T2}. (left) box trajectory; (right) cost density.}
    \label{fig:T2}
\end{figure}

\paragraph{T2. Circular-path pushing:} Here, we verify that this performance holds under a continuously changing pushing direction. To verify that closed-loop performance holds under a continuously changing pushing direction, we evaluate circular-path pushing. Figure~\ref{fig:T2} compares the three control modes: \emph{True} and \emph{Adaptive-MHE} closely track the reference path (left), while \emph{Baseline} drifts and fails to maintain the desired radius. The cost distributions (right) are consistent with this trend: \emph{Adaptive-MHE} shows a sharp peak near zero cost, indicating efficient pushes, whereas \emph{Baseline}'s broader, higher-cost distribution reflects degraded performance from the mismatched dynamics.

\begin{figure}[htbp]
\centering
\includegraphics[width=.8\linewidth, trim={0.0cm 4.5cm 0.0cm 5.2cm}, clip]{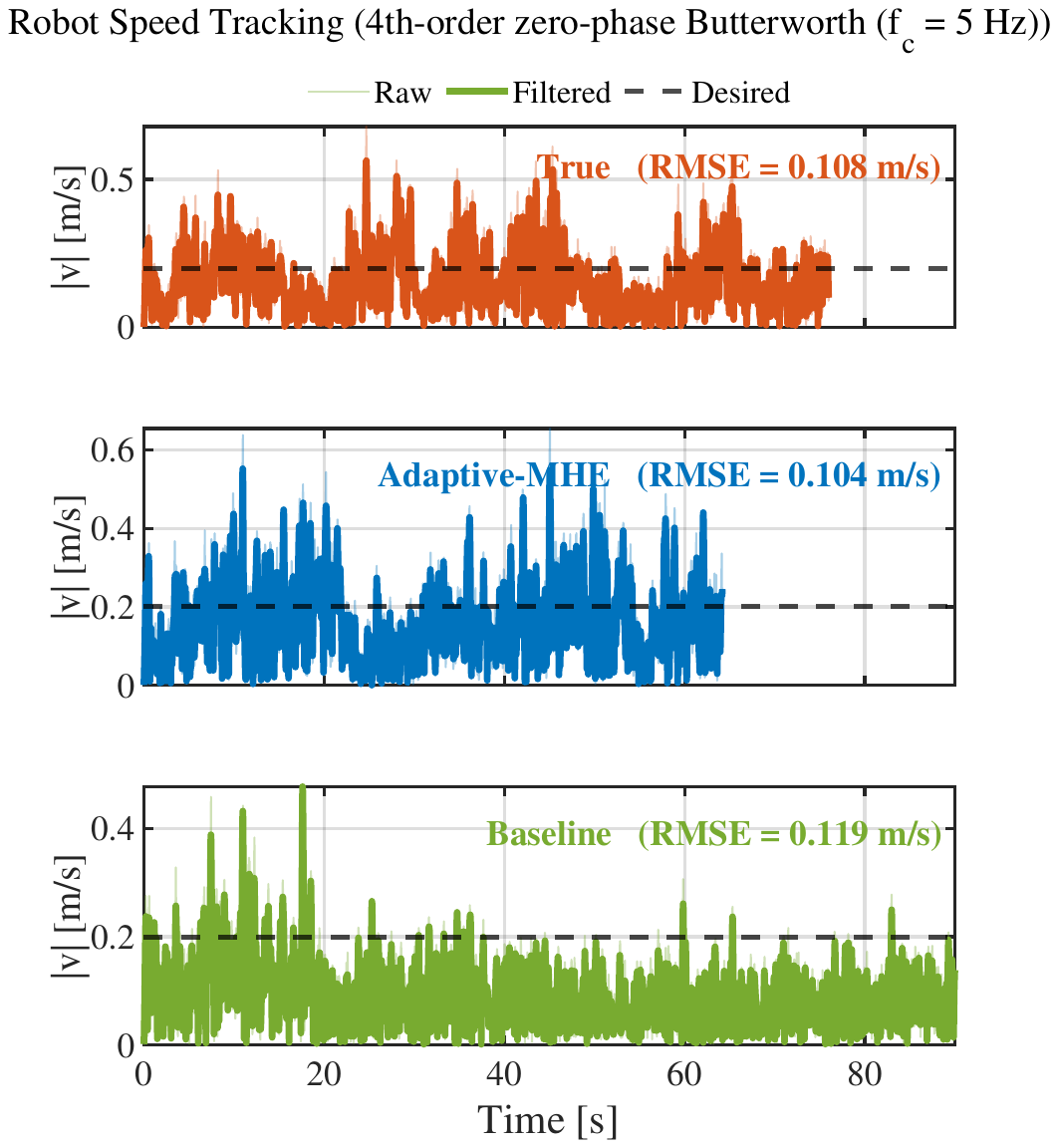}
\caption{Robot forward velocity tracking for \hyperref[task:T2]{T2}.}
\label{fig:velocity_circular}
\end{figure}

Figure~\ref{fig:velocity_circular} shows the robot's speed against the desired $0.2$~m/s target. \emph{True} and \emph{Adaptive-MHE} track the desired speed closely with lower RMSE than \emph{Baseline}, confirming that even under a continuously changing pushing direction, \emph{Adaptive-MHE} matches \emph{True} and consistently outperforms \emph{Baseline}.

\begin{figure}[htbp]
	\centering
    \begin{subfigure}[]{0.43\linewidth}
	\includegraphics[width=\linewidth, trim={2.0cm 7.0cm 4.5cm 7.5cm}, clip]{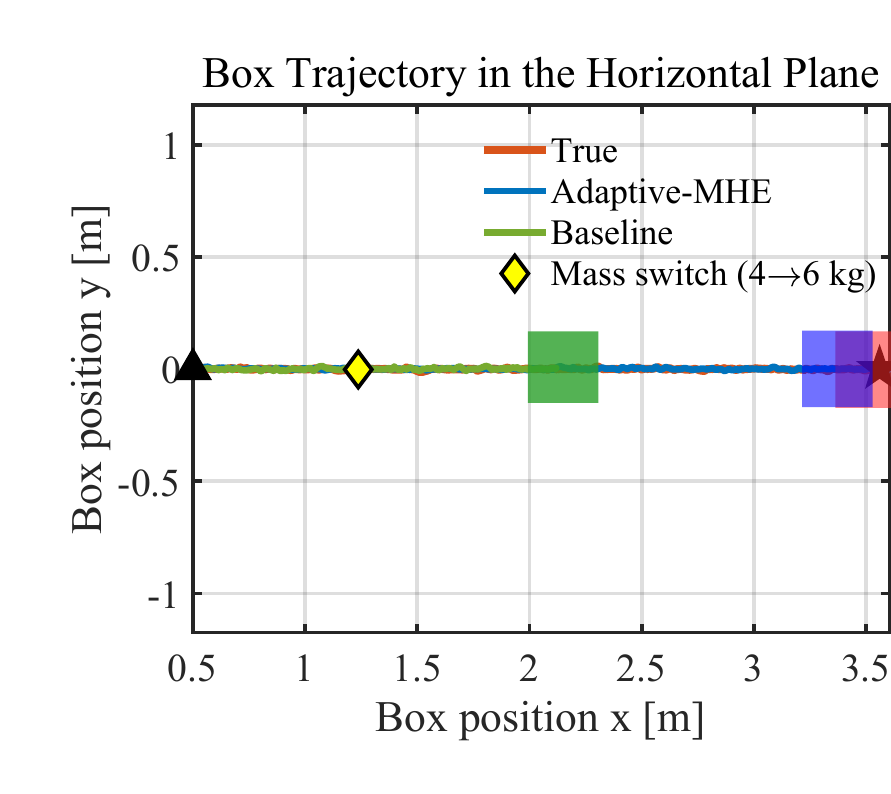}
	\label{fig:mass_switch_traj}
    \end{subfigure}%
    \hfill
    \begin{subfigure}[]{0.57\linewidth}
    \includegraphics[width=\linewidth, trim={0.0cm 7.0cm 0.0cm 7.0cm}, clip]{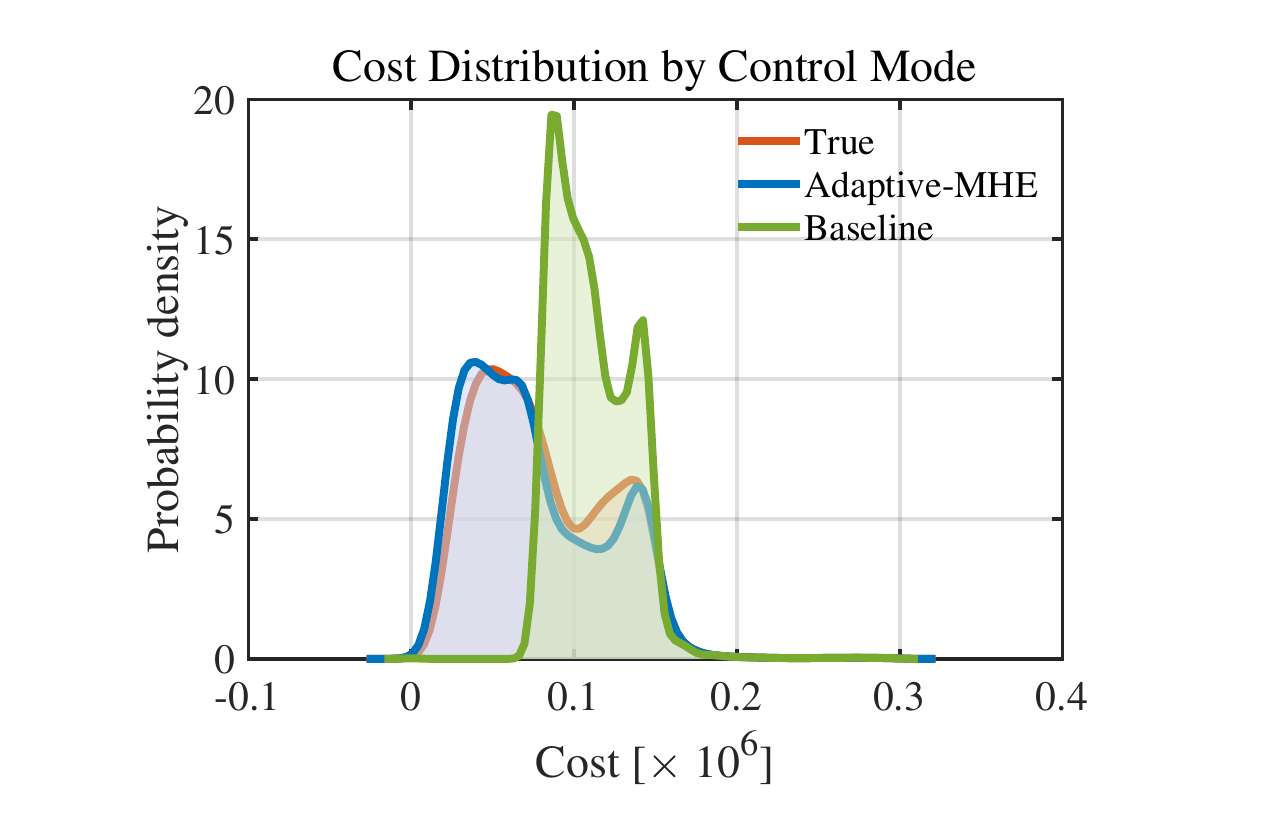}
	\label{fig:mass_switch_cost}
    \end{subfigure}%
    \vspace{-5mm}
    \caption{Comparison of control modes for T3. (left) box trajectory; (right) cost density.}
    \label{fig:T3_traj}
    \vspace{-1mm}
\end{figure}

\paragraph{T3. Adaptation to mass variation:} We next test robustness to a sudden change in the object being pushed. Figure~\ref{fig:T3_traj} compares the three control modes when the box mass is switched from $4$ to $6$~kg at $t=10$~s. \emph{True} and \emph{Adaptive-MHE} continue tracking the reference path closely after the switch (right), while \emph{Baseline} deviates noticeably, unable to account for the updated mass; the corresponding cost distributions show the same pattern.

\begin{figure}[htbp]
	\centering
    \begin{subfigure}[]{0.46\linewidth}
	\includegraphics[width=\linewidth, trim={2.0cm 5.0cm 3.0cm 4.5cm}, clip]{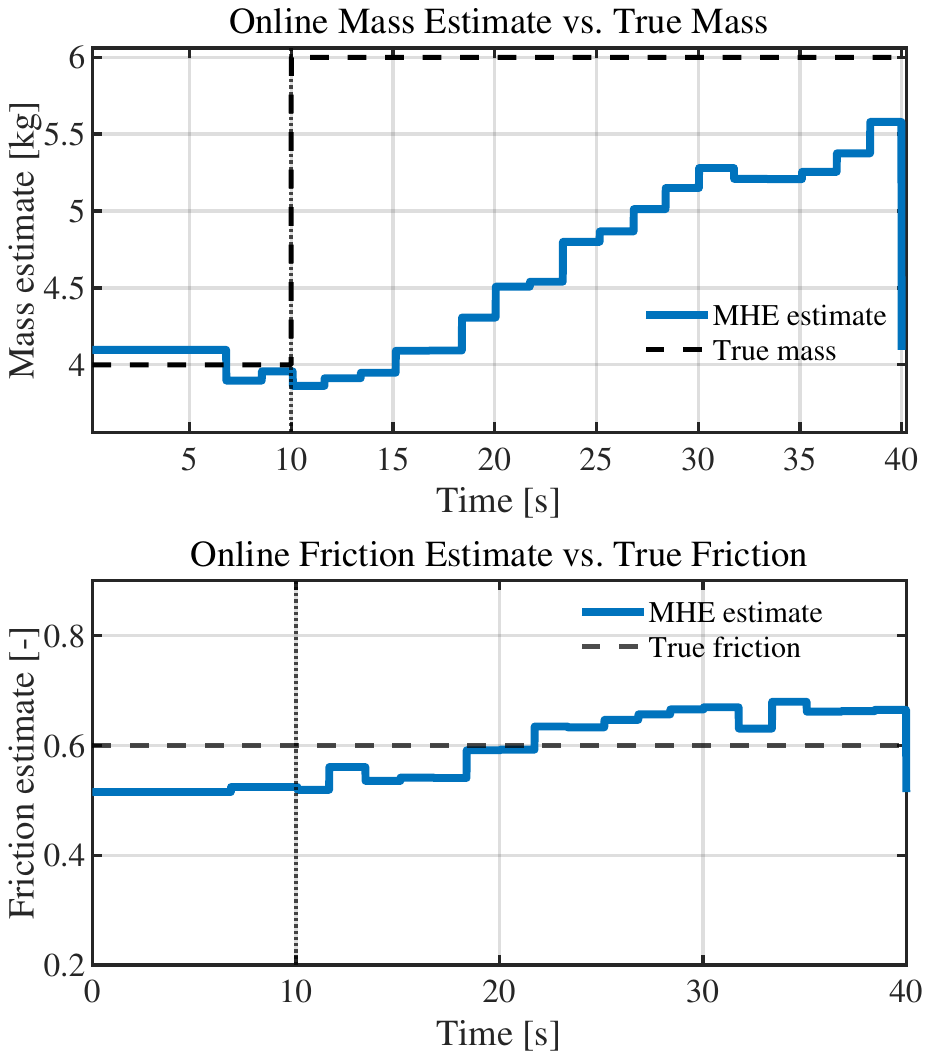}
	\label{fig:mass_var_est}
    \end{subfigure}%
    \hfill
    \begin{subfigure}[]{0.54\linewidth}
    \includegraphics[width=\linewidth, trim={1.0cm 4.0cm 1.0cm 5.0cm}, clip]{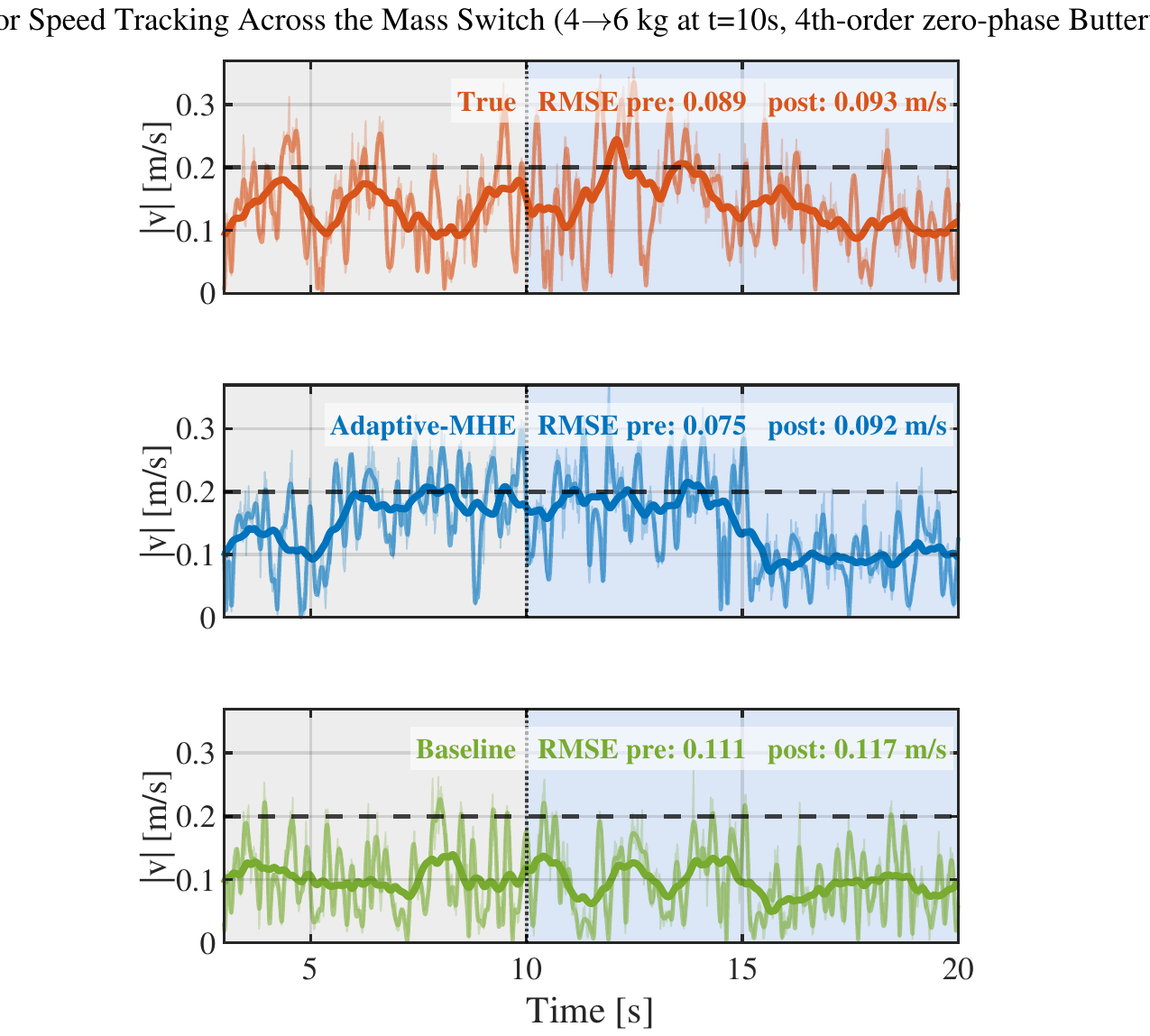}
	\label{fig:mass_var_vel}
    \end{subfigure}%
    \vspace{-5mm}
    \caption{Comparison of control modes for T3. (left) estimated parameters, (right) robot forward velocity tracking.}
    \label{fig:T3_vel}
\end{figure}

Figure~\ref{fig:T3_vel} illustrates the underlying online adaptation: \emph{Adaptive-MHE} tracks the true mass and friction coefficient closely, updating its estimates once the switch occurs and converging shortly after (left). Speed tracking (right) shows \emph{True} and \emph{Adaptive-MHE} maintain low error immediately after the switch, while \emph{Baseline} shows a substantially larger RMSE of $0.117$~m/s (Table~\ref{tab:sim_summary}), reflecting its inability to adapt. These results confirm the proposed framework detects and compensates for abrupt, unmodeled changes in object properties online.

\begin{figure}[htbp]
	\centering
    \begin{subfigure}[]{0.46\linewidth}
	\includegraphics[width=\linewidth, trim={2.0cm 5.0cm 3.0cm 4.5cm}, clip]{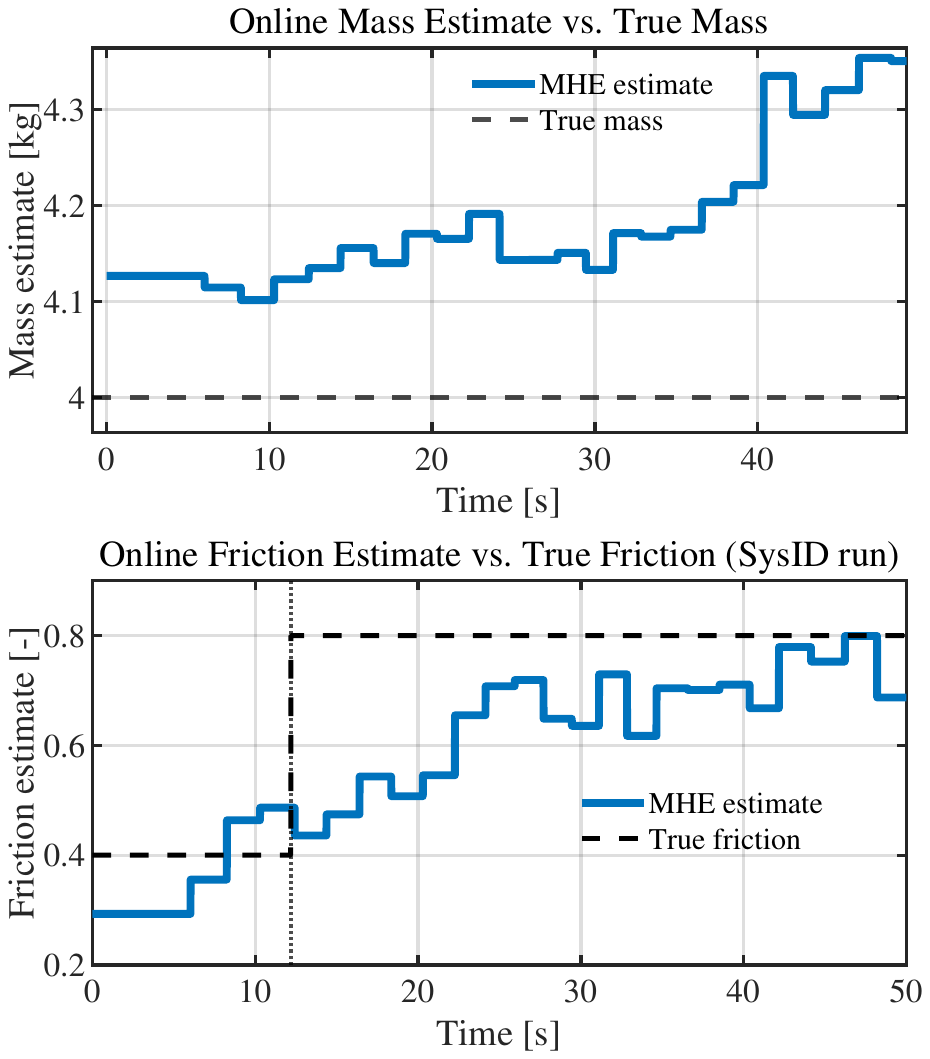}
	\label{fig:params_est}
    \end{subfigure}%
    \hfill
    \begin{subfigure}[]{0.54\linewidth}
    \includegraphics[width=\linewidth, trim={1.0cm 4.0cm 1.0cm 5.0cm}, clip]{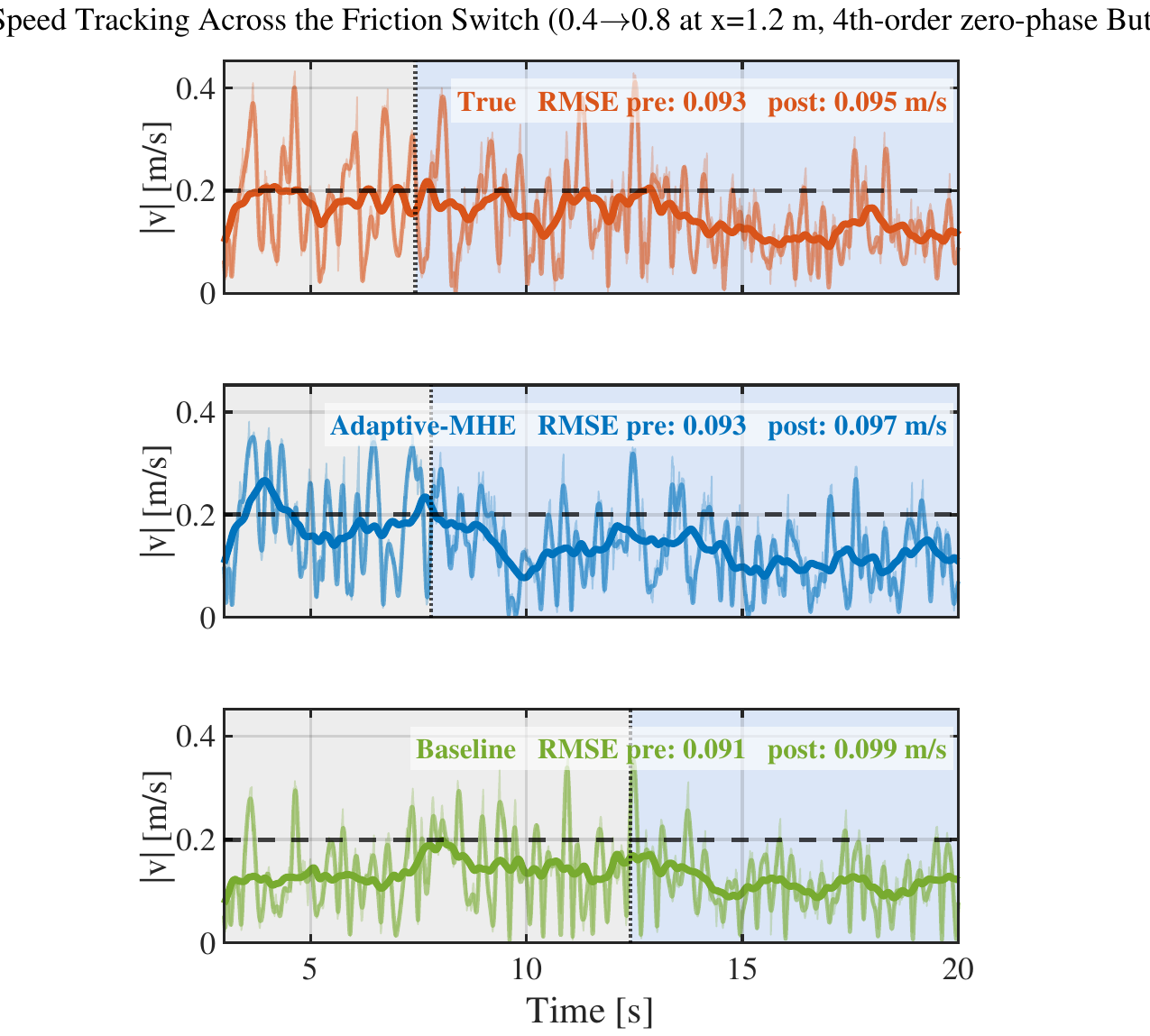}
	\label{fig:fric_var_vel}
    \end{subfigure}%
    \vspace{-5mm}
    \caption{Comparison of control modes for T4. (left) estimated parameters, (right) robot forward velocity tracking.}
    \label{fig:T4}
\end{figure}

\paragraph{T4. Adaptation to terrain change:} We next turn to a changing environment, i.e., the robot traverses a floor with an abrupt change in surface friction. As shown in Fig.~\ref{fig:T4} (left), \emph{Adaptive-MHE}'s mass estimate remains close to its true constant value ($m_{\text{true}}=4$~kg) throughout the run, with only a brief transient at the switch, while the friction estimate converges toward the true post-switch value within a few seconds of crossing $x_b=1.5$~m.

Figure~\ref{fig:T4} (right) shows the resulting effect on closed-loop speed tracking. Since the switch triggers on box position, \emph{True} and \emph{Adaptive-MHE} cross $x_b=1.5$~m sooner than \emph{Baseline}, which pushes more slowly; pre- and post-switch windows therefore span different absolute intervals across modes. All three achieve comparable pre-switch accuracy, with \emph{Baseline} only marginally worse. Post-switch, tracking degrades modestly for all modes, with \emph{Adaptive-MHE}'s degradation mirroring the brief re-convergence transient in Fig.~\ref{fig:T4} (left), suggesting the benefit of online friction adaptation is not strongly separable from the fixed-parameter controller at the level of closed-loop velocity tracking alone, even though the underlying friction estimate is substantially more accurate under \emph{Adaptive-MHE}.

\begin{figure}[htbp]
\centering
\includegraphics[width=0.9\linewidth, trim={2.0cm 19.5cm 2.0cm 2.0cm}, clip]{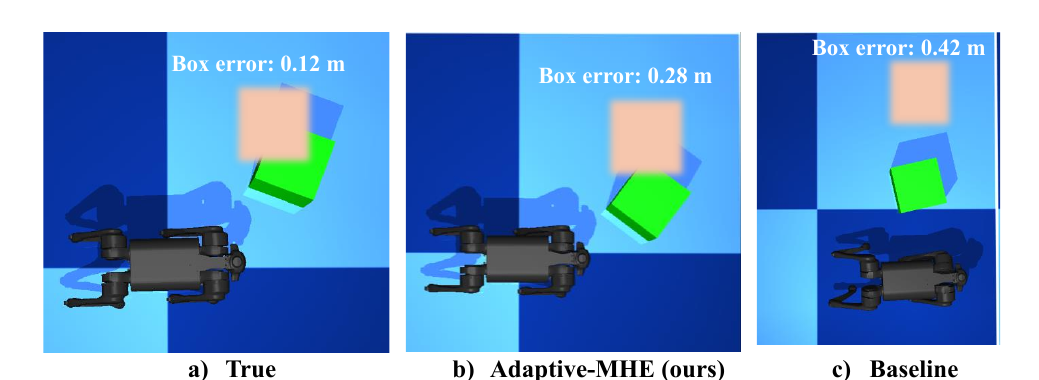}
\caption{Final configuration of the robot and box after completing \hyperref[task:T5]{T5}.}
\label{fig:push_one_leg_results}
\end{figure}

\begin{figure}[htbp]
\centering
\includegraphics[width=0.9\linewidth, trim={0.0cm 8.8cm 0.0cm 7.5cm}, clip]{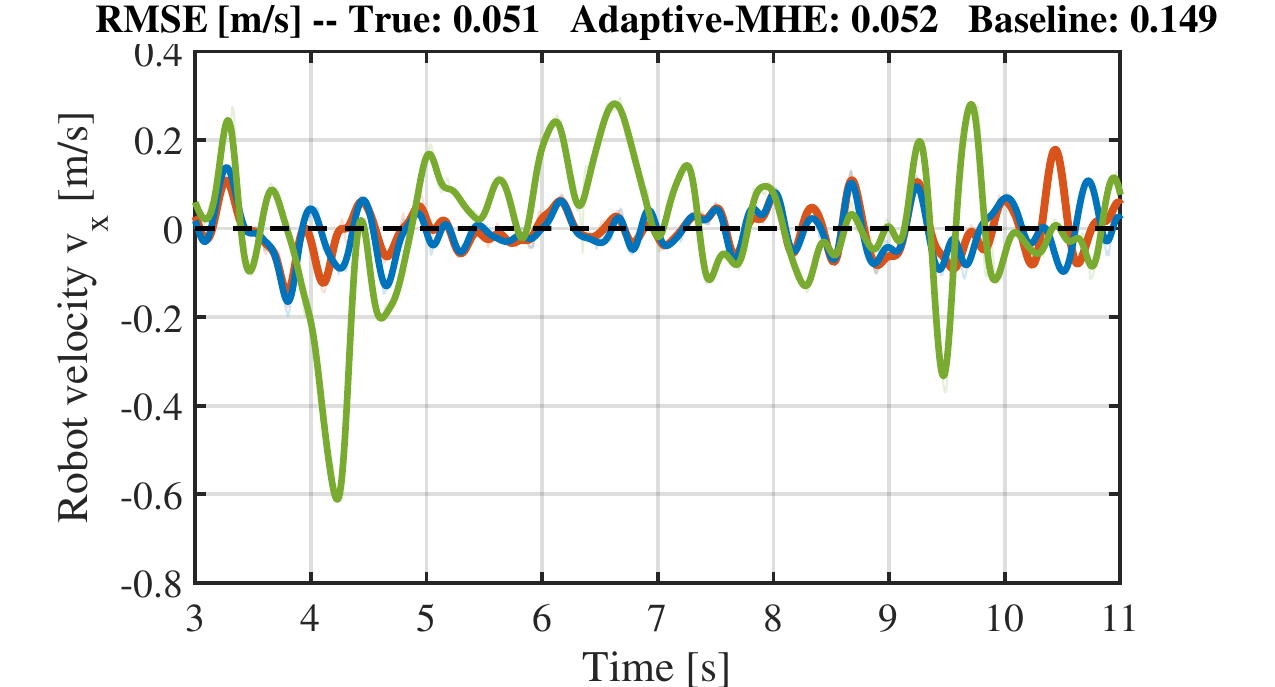}
\caption{Robot speed tracking for \hyperref[task:T5]{T5}.}
\label{fig:velocity_one_leg}
\end{figure}

\paragraph{T5. Pushing a box with one leg:} Finally, we test the framework beyond whole-body pushing by having the robot manipulate the box with a single leg (Fig.~\ref{fig:tasks}.e). Figure~\ref{fig:push_one_leg_results} compares the final box configuration for a $0.5$~m lateral push: \emph{Adaptive-MHE} achieves a final box error of $0.28$~m, closely approaching \emph{True} ($0.12$~m), while \emph{Baseline} incurs a substantially larger error of $0.42$~m. Also, Fig.~\ref{fig:velocity_one_leg} shows \emph{True} and \emph{Adaptive-MHE} closely track the zero-velocity reference (RMSE $0.051$ and $0.052$~m/s), while \emph{Baseline} shows a markedly higher RMSE of $0.149$~m/s. These results demonstrate that single-leg pushing, paired with online adaptation, achieves accurate object displacement despite the added complexity of using a locomotion limb as a temporary manipulator.


\end{document}